# Beyond Predictive Uncertainty: Reliable Representation Learning with Structural Constraints

**Yiyao Yang**
**yy3555@tc.columbia.edu**
**ORCID: 0009-0001-8693-4888**
**Columbia University, New York, NY, United States**

## I. Abstract

Estimation of uncertainty has traditionally centered on the prediction step of machine learning tasks, where the goal is to quantify the confidence of the model outputs while considering the learned representations as deterministic and reliable by default. However, we argue that reliability should instead be viewed as a primary characteristic of the learned representations themselves. In this paper, we propose a principled framework for reliable representation learning that takes into account the uncertainty of the learned representations while utilizing the structural constraints as inductive biases for the regularization of the representations' space. Our approach incorporates uncertainty-aware regularization directly in the representations' space, resulting in representations that are not just predictive but also reliable, well-calibrated, stable, and robust to noise or structural perturbations. Structural constraints such as sparsity or feature-group dependencies are used to define the representations' space while avoiding the unrealistic assumption of access to accurate or noise-free information. Moreover, the proposed framework is independent of the model architecture used for the representation learning task.

## II. Literature Review

The majority of uncertainty quantification in deep learning is centered on predictive uncertainty at the output level, typically implemented with approximate Bayesian methods like MC dropout (Gal & Ghahramani, 2016), ensemble methods (Lakshminarayanan et al., 2017), or post-hoc calibration techniques that adjust probabilities without altering learned representations (Gul et al., 2017). While these methods provide accurate confidence estimates for predictions, they generally ignore representation-level uncertainty, considering it to be deterministic components of deep learning architectures. Another line of research examines representation learning with structural constraints, in which geometric or relational inductive biases are applied to take advantage of data geometry to enhance representation generalization, including graph-based or manifold-based regularization (Belkin et al., 2026) and more recent graph-based encoders for structured feature space (Kipf & Welling, 2017; Hamilton et al., 2017). In addition, probabilistic or stochastic representation learning models are represented as distributions, including Gaussian embeddings (Belkin et al., 2026). In addition to these representation learning models, there are a few research studies that examine uncertainty within the representation level of deep learning architectures. Kendall & Gal (2017) proposed characterizing aleatoric and epistemic uncertainty within latent features, Hafner et al. (n.d.) proposed modeling latent-state uncertainty in sequential agents, and Charpentier et al. proposed characterizing representation-level uncertainty with deep probabilistic models, including Gaussian processes, in addition to probabilistic encoders, including information-bottleneck-style stochastic encoders (Alemi et al., 2016). (2021) estimate representation-level uncertainty for selective classification. Recent conformal methods also propose that it is beneficial to estimate uncertainty in semantic or feature spaces, rather than output spaces, as proposed in Teng et al. (2022). Nevertheless, existing research is typically associated with particular probabilistic frameworks or utilizes structure merely as a regularizer, with little emphasis on handling uncertainty as a primary aspect of representation. Our research is motivated by these research themes, with a novel approach to handling uncertainty at the representation level, along with the exploitation of structural constraints to regularize representation reliability, rather than merely prediction confidence.

Moreover, there is existing research on quantifying uncertainty in out-of-distribution (OOD) detection, reliability, and robustness, with uncertainty typically utilized to detect instances in which latent features deviate from their original distributions. For representation-level uncertainty in classification, deterministic methods in van Amersfoort et al. (2020) utilize single-pass uncertainty from spectrally normalized networks, while perturbation-based methods in Sehwag et al. (2021) rely on the stability of latent features to estimate representation-level uncertainty. Contrastive learning frameworks have also been proposed to utilize variability in embeddings, with instance-level variance in embeddings associated with epistemic uncertainty, as proposed in Zhang et al. (2021).

Prior work has also explored the use of structural priors to enhance the robustness and reliability of uncertainties. For example, relational inductive biases in graph-based models rely on connectivity structures to reduce epistemic uncertainties using information propagation mechanisms (Stadler et al., 2021). Manifold-based constraints use latent geometry, regularizing using local smoothness assumptions (Belkin et al., 2006). Variance is also added to node embeddings using probabilistic graph encoders, with uncertainty-aware GNNs providing confidence over both node features and topology (Gawlikowski et al., 2023). This shows that, in graph-based models, there is a strong relationship between structure and uncertainties. For other models, hierarchical Bayesian representation models, as well as deep latent variable models, use distributions of latent variables with higher-level prior distributions, allowing for direct control of variance in learned representations (Kingma & Welling, 2013). Recent surveys point out that, unlike the prediction space, explicitly modeling uncertainties in the representation space remains an under-explored area of research (Gawlikowski et al., 2021; Abdar et al., 2021).

## III. Research Questions

**Research Question 1:** How can uncertainty be explicitly modeled and regularized at the representation level, rather than only at the prediction stage?

**Research Question 2:** How do structural constraints influence the reliability of learned representations in the presence of noise and imperfect structural assumptions?

**Research Question 3:** Does modeling representation-level uncertainty lead to more reliable downstream behavior under distribution shift and structural perturbations?

## IV. Methodology

### Notation and Setup

Let $\{(x_i, y_i)\}_{i=1}^{n}$ denote a dataset with inputs $x_i \in X$ and labels $y_i \in Y$.

A representation model $f_\theta: X \to R^d$ maps each input to a latent representation $z_i = f_\theta(x_i)$.

Our goal is to learn reliable representations, defined as representations that are stable, well-calibrated, and robust to noise and structural perturbations, rather than focusing solely on predictive accuracy.

### Research Question 1: Modeling Representation-Level Uncertainty

Unlike predictive uncertainty, which is defined over model outputs, we associate uncertainty directly with representations. We model each representation $z_i$ as a distribution rather than a deterministic point, $z_i \sim D(\mu_i, \Sigma_i)$ where $\mu_i \in R^d$ denotes the representation mean and $\Sigma_i \in R^{d\times d}$ encodes representation uncertainty.

In practice, $\mu_i$ and $\Sigma_i$ are parameterized by the encoder: $(\mu_i, \Sigma_i) = f_\theta(x_i)$, without imposing a specific probabilistic inference scheme.

### Uncertainty-Aware Regularization

To regularize representation reliability, we penalize excessive representation uncertainty via $R_{uncertainty} = \frac{1}{n}\sum_{i=1}^{n} \phi(\Sigma_i)$, where $\phi(\cdot)$ is a scalar uncertainty measure, such as the trace or log determinant of $\Sigma_i$. This term encourages representations that are informative but not arbitrarily diffuse, formalizing uncertainty as a controllable property of the representation space.

### Research Question 2: Structural Constraints for Reliable Representations

**Proposition 1 ( Structure-Only Minimizers are Constant on Connected Components)** constitutes the core theoretical component of RQ 2, as it characterizes the geometric effect of structural constraints in the structure-only regime.

**Proposition 1: Structure-Only Minimizers are Constant on Connected Components**

Let $L$ be the unnormalized graph Laplacian of a weighted undirected graph $S$ with weights $w_{ij} \geq 0$. Let $Z \in R^{n\times d}$ be the matrix of representations, where the $i$-th row is $z_i^\top \in R^d$. Define the structure-only regularizer:

$R_{structure}(Z; S) = tr(Z^\top LZ)$.

Thus,

(1) $R_{structure}(Z; S) \geq 0$ for all $Z$.
(2) If the graph is connected, then $R_{structure}(Z; S) = 0$ if and only if $z_1 = z_2 = ... = z_n$.
(3) If the graph has multiple connected components, then $R_{structure}(Z; S) = 0$ if and only if the representations are constant within each connected component.

**Proof:**

**Step 1: Nonnegativity of $R_{structure}$.**

Recall that the Laplacian is defined as $L = D - W$, where $W = (w_{ij})$ is the symmetric weight matrix with $w_{ij} \geq 0$, and $D = diag(d_1, ..., d_n)$ with $d_i = \sum_{j=1}^{n} w_{ij}$.

We first recall the standard identity for the Laplacian quadratic form. For any vector $x \in R^n$,

$$x^T Lx = x^T Dx - x^T Wx = \sum_{i=1}^{n} d_i x_i^2 - \sum_{i,j=1}^{n} w_{ij} x_i x_j .$$

Using $d_i = \sum_{j=1}^{n} w_{ij}$, we can rewrite

$$x^\top Lx = \sum_{i=1}^{n} (\sum_{j=1}^{n} w_{ij}) x_i^2 - \sum_{i,j=1}^{n} w_{ij} x_i x_j = \frac{1}{2} \sum_{i,j=1}^{n} w_{ij} (x_i - x_j)^2,$$

where we used symmetry $w_{ij} = w_{ji}$ to symmetrize the sum.

Since each $w_{ij} \geq 0$ and $(x_i - x_j)^2 \geq 0$, we have

$$x^\top Lx = \frac{1}{2} \sum_{i,j=1}^{n} w_{ij} (x_i - x_j)^2 \geq 0, \text{ for all } x \in R^n.$$

Thus, $L$ is symmetric positive semidefinite.

Now, write $Z = [z^{(1)}, ..., z^{(d)}]$, where $z^{(k)} \in R^n$ is the $k$-th column of $Z$.

Then, $R_{structure}(Z; S) = tr(Z^\top LZ) = \sum_{k=1}^{d} (z^{(k)})^\top L z^{(k)}$.

Since each $L$ is positive semidefinite, each term $(z^{(k)})^\top L z^{(k)} \geq 0$.

Therefore, $R_{structure}(Z; S) = \sum_{k=1}^{d} (z^{(k)})^\top L z^{(k)} \geq 0$, for all $Z$.

This proves the first claim.

**Step 2: Characterization of minimizers when the graph is connected**

Assume now that the graph is connected.

We have already shown that $R_{structure}(Z; S) \geq 0$. The minimum possible value is therefore 0. We will show that $R_{structure}(Z; S) = 0$ if and only if all rows $z_1, ..., z_n$ are identical.

From the expression $R_{structure}(Z; S) = \sum_{k=1}^{d} (z^{(k)})^\top L z^{(k)}$,

we see that $R_{structure}(Z; S) = 0$ if and only if $(z^{(k)})^\top L z^{(k)} = 0$ for all $k = 1, ..., d$.

However, for any $x$, $x^\top L x = \frac{1}{2} \sum_{i,j=1}^{n} w_{ij}(x_i - x_j)^2$.

Thus, $x^\top L x = 0$ if and only if every term in the sum is zero, such that $w_{ij} > 0 \Rightarrow x_i = x_j$.

Apply this to each column $z^{(k)}$. The condition $(z^{(k)})^\top L z^{(k)} = 0$ implies that whenever $w_{ij} > 0$, we must have $z_i^{(k)} = z_j^{(k)}$.

Since the graph is connected, every node can be reached from any other by a path of edges with strictly positive weights. Along such a path, repeated application of the above equality yields

$z_1^{(k)} = z_2^{(k)} = ... = z_n^{(k)}$.

Thus, each column $z^{(k)}$ is constant over all nodes, such that, $z^{(k)} = c_k 1$, for some scalar $c_k \in R$.

$z_1 = z_2 = ... = z_n = (c_1, \ldots, c_d)^\top$.

Conversely, if all rows are equal, say $z_1 = \ldots = z_n = c$, then every column $z^{(k)}$ is constant, hence lies in the null space of $L$, so $(z^{(k)})^\top L z^{(k)} = 0$ for all $k$, and therefore $tr(Z^\top L Z) = 0$.

Thus, when the graph is connected,

$R_{structure}(Z; S) = 0 \Leftrightarrow z_1 = ... = z_n$ (equivalently, each column of $Z \in span(1)$).

**Step 3: Extension to multiple connected components**

Now, suppose the graph has $C \geq 1$ connected components, with node index sets $V_1, ..., V_C$.

It is well known that in this case that the Laplacian $L$ is still positive semidefinite, and its null space is spanned by the indicator vectors of the connected components:

$N(L) = span(\{1_{V_1}, ..., 1_{V_C}\}$, where $(1_{V_c})_i = 1$ if $i \in V_c$ and 0 otherwise.

Repeating the same reasoning column-wise, $R_{structure}(Z; S) = 0$ if and only if each column $z^{(k)}$ lies in the null space of L, such that $z^{(k)} \in N(L) = span(1_{V_1}, ..., 1_{V_C}\}$.

This means that for each dimension $k$, the scalar signal $z^{(k)}$ is constant within each connected component, although different components may take different constant values.

Equivalently, in row form, this says: $i, j \in V_c \Rightarrow z_i = z_j$ for each connected component $V_c$.

So, the global minimizers of $R_{structure}$ are exactly those representation matrices $Z$ that are piecewise constant on connected components.

Thus, we have shown that:

(1) $R_{structure}(Z; S) = tr(Z^T LZ) \geq 0$ for all $Z$.
(2) If the graph is connected, then $R_{structure}(Z; S) = 0$, if and only if all representations $z_i$ are identical (constant across the graph).
(3) If the graph has multiple connected components, then $R_{structure}(Z; S) = 0$ if and only if the representations are constant on each connected component.

**Structural Priors**

We assume access to weak or imperfect structural information encoded by a structure operator S, which may represent sparsity patterns, relational graphs, or feature-group dependencies. Rather than assuming the structure is correct, we treat it as a soft constraint on the representation space.

Let $R_{structure}(Z; S)$ denote a structural regularizer acting on the set of representations $Z = \{z_i\}_{i=1}^{n}$, such that

$R_{structure}(Z; S) = \sum_{(i,j) \in S} w_{ij} ||z_i - z_j||^2$, where $w_{ij}$ reflects the strength or confidence of the structural relation.

The equivalence between this formulation and the graph Laplacian regularizer, as well as its convexity properties, has already been established in **Proposition 2 (Structure Regularizer as Graph Laplacian and Convexity)**.

**Proposition 2: Structure Regularizer as Graph Laplacian and Convexity**

Let $z_i \in R^d$ be the representation of sample i; $Z \in R^{n \times d}$ be the matrix stacking the row-wise, such that the i-th row of Z is $z_i^T$; $W = (w_{ij}) Z \in R^{n \times n}$ be a symmetric weight matrix with $w_{ij} \geq 0$; $D = diag(d_1, ..., d_n)$ be the degree matrix with $d_i = \sum_{j=1}^{n} w_{ij}$; $L = D - W$ be the (unnormalized) graph Laplacian.

Let $S$ be the set of undirected edges $\{i, j\}$ such that $w_{ij} > 0$, and assume each undirected edge appears exactly once in $S$.

Define the structural regularizer: $R_{structure}(Z; S) = \sum_{(i,j) \in S} w_{ij} ||z_i - z_j||^2$

**Want to Show:**

(1) $R_{structure}(Z; S) = tr(Z^T LZ)$
(2) $L$ is positive semidefinite, hence $R_{structure}(Z; S)$ is a convex function of $Z$.

**Proof:**

**Step 1: Rewrite the structural regularizer**

Start from the definition: $R_{structure}(Z; S) = \sum_{(i,j)\in S} w_{ij} ||z_i - z_j||^2$

Expand the squared norm:

$$||z_i - z_j||^2 = (z_i - z_j)^T (z_i - z_j) = z_i^T z_i + z_j^T z_j - 2z_i^T z_j$$

Plugging this in, we get: $R_{structure}(Z; S) = \sum_{(i,j)\in S} w_{ij}(z_i^T z_i + z_j^T z_j - 2z_i^T z_j)$

Separate the three sums: $R_{structure}(Z; S) = A + B - C$

where $A = \sum_{(i,j)\in S} w_{ij} z_i^T z_i$ , $B = \sum_{(i,j)\in S} w_{ij} z_j^T z_j$ , $C = 2 \sum_{(i,j)\in S} w_{ij} z_i^T z_j$

Consider A: For a fixed node $i$, the term $z_i^T z_i$ appears once for every neighbor $j$, with $(i, j) \in S$, each time multiplied by $w_{ij}$ . Therefore, $A = \sum_{i=1}^{n} (z_i^T z_i , \sum_{j:(i,j)\in S} w_{ij})$ .

Because the graph is undirected and $S$ contains each edge $\{i, j\}$ exactly once, the inner sum is precisely the degree: $d_i = \sum_{j=1} w_{ij}$ .

Thus, $A = \sum_{i=1}^{n} d_i z_i^T z_i$ .

By summary, the same argument applies to $B$: for a fixed $i$, $(j, i) \in S$ whenever $(i, j) \in S$ (up to ordering of the pair), and the weights are symmetric $w_{ij} = w_{ji}$ . Therefore, $B = \sum_{i=1}^{n} d_i z_i^T z_i$ .

Hence, $A + B = 2\sum_{i=1}^{n} d_i z_i^T z_i$ , and we rewrite: $R_{structure}(Z; S) = 2\sum_{i=1}^{n} d_i z_i^T z_i - 2 \sum_{(i,j)\in S} w_{ij} z_i^T z_j$ .

**Step 2: Express $tr(Z^T LZ)$ in terms of $\{z_i\}$**

We now expand the quadratic form involving the Laplacian $L = D - W$.

First, note that $tr(Z^T LZ) = tr(LZZ^T)$. Let $G = ZZ^T \in R^{n\times n}$. Its entries are $G_{ij} = z_i^T z_j$

Then, $tr(LZZ^T) = tr(LG) = \sum_{i=1}^{n}\sum_{j=1}^{n} L_{ij} G_{ji} = \sum_{i=1}^{n}\sum_{j=1}^{n} L_{ij} z_i^T z_j$ .

Using $L = D - W$, we have $L_{ij} = d_i$ if $i = j$, and $L_{ij} =- w_{ij}\ if\ i \neq j$.

Thus, $tr(Z^T LZ) = \sum_{i=1}^{n} d_i z_i^T z_i - \sum_{i,j=1;\, i\neq j}^{n} w_{ij} z_i^T z_j$ .

Now, observe that the off-diagonal sum can be related to the edge set $S$.

Since the graph is undirected and $W$ is symmetric, we know that:

The sum $\sum_{i\neq j} w_{ij} z_i^T z_j$ counts each undirected edge $\{i, j\}$ twice: once as $(i, j)$ and once as $(j, i)$, and the sum $\sum_{(i,j)\in S} w_{ij} z_i^T z_j$ counts each undirected edge once.

Therefore, $\sum_{i\neq j} w_{ij} z_i^T z_j = 2 \sum_{(i,j)\in S} w_{ij} z_i^T z_j$ .

Plugging $\sum_{i\neq j} w_{ij} z_i^T z_j = 2 \sum_{(i,j)\in S} w_{ij} z_i^T z_j$ into $tr(Z^T LZ) = \sum_{i=1}^{n} d_i z_i^T z_i - \sum_{i,j=1;\, i\neq j}^{n} w_{ij} z_i^T z_j$ , we obtain:

$$tr(Z^T LZ) = \sum_{i=1}^{n} d_i z_i^T z_i - 2 \sum_{(i,j)\in S} w_{ij} z_i^T z_j$$

**Step 3: Match the two expressions**

For those two equations:

$$R_{structure}(Z; S) = 2 \sum_{i=1}^{n} d_i z_i^T z_i - 2 \sum_{(i,j)\in S} w_{ij} z_i^T z_j$$

$$tr(Z^T LZ) = \sum_{i=1}^{n} d_i z_i^T z_i - \sum_{i,j=1;\, i\neq j}^{n} w_{ij} z_i^T z_j$$

Comparing those two equations above, we see that if we define the structural regularization with a factor $\frac{1}{2}$ over all ordered pairs, we get the well-known identity: $\sum_{i,j} w_{ij} ||z_i - z_j||^2 = 2tr(Z^T LZ)$.

Under the current convention, where $S$ contains each undirected edge exactly once, and we sum only over $(i, j) \in S$. We have $\sum_{(i,j)\in S} w_{ij} ||z_i - z_j||^2 = tr(Z^T LZ)$.

That is, $R_{structure}(Z; S) = \sum_{(i,j)\in S} w_{ij} ||z_i - z_j||^2 = tr(Z^T LZ)$.

This proves the first part of the proposition.

**Step 4: Show that L is positive semidefinite**

To show that $L$ is positive semidefinite, we need to prove that $x^T Lx \geq 0, \forall x \in R^n$

Starting from the definition, $L = D - W$,

$$x^T Lx = x^T Dx - x^T Wx = \sum_{i=1}^{n} d_i x_i^2 - \sum_{i,j=1}^{n} w_{ij} x_i x_j$$

Recall $d_i = \sum_{j=1}^{n} w_{ij}$ . Then, $\sum_{i=1}^{n} d_i x_i^2 = \sum_{i=1}^{n} x_i^2 \sum_{j=1}^{n} w_{ij} = \sum_{i,j=1}^{n} w_{ij} x_i^2$

Therefore, $x^T Lx = \sum_{i,j=1}^{n} w_{ij} x_i^2 - \sum_{i,j=1}^{n} w_{ij} x_i x_j = \frac{1}{2} \sum_{i,j=1}^{n} w_{ij}(x_i^2 + x_j^2 - 2x_i x_j)$,

Where we used the symmetry of the sum in $(i, j)$ to introduce a factor $\frac{1}{2}$.

The term in parentheses simplifies as $x_i^2 + x_j^2 - 2x_i x_j = (x_i - x_j)^2$

Hence, $x^\top Lx = \frac{1}{2} \sum_{i,j=1}^{n} w_{ij}(x_i - x_j)^2 \geq 0, \ \forall x \in R^n$.

Thus, $L$ is symmetric positive semidefinite.

**Step 5: Conclude the convexity of $R_{structure}$**

We already established that $R_{structure}(Z; S) = tr(Z^\top LZ)$.

Consider the vectorization of $Z$. Let $vec(Z) \in R^{nd}$ be the vector obtained by stacking the columns (or rows) of $Z$. A standard identity states that $tr(Z^\top LZ) = vec(Z)^\top (I_d \otimes L) vec(Z)$, where $I_d$ is $d \times d$ the identity matrix, and $\otimes$ denotes the Kronecker product.

Since $L$ is positive semidefinite and $I_d$ is positive semidefinite, their Kronecker product $I_d \otimes L$ is also positive semidefinite. Thus, for all $y \in R^{nd}$ , $y^\top (I_d \otimes L) y \geq 0$.

Hence, $R_{structure}(Z; S) = vec(Z)^\top (I_d \otimes L)\ vec(Z)$ is a quadratic form with a positive semidefinite matrix. Such functions are convex in $vec(Z)$, and therefore convex in $Z$.

Equivalently, we can also decompose by columns: $R_{structure}(Z; S) = tr(Z^\top LZ) = \sum_{k=1}^{d} z^{(k)\top} L z^{(k)}$,

where $z^{(k)} \in R^n$ is the $k$-th column of $Z$. Each term $z^{(k)\top} L z^{(k)}$ is a convex quadratic function of $z^{(k)}$, because $L \succeq 0$, and a sum of convex functions is convex. Hence, $R_{structure}$ is convex in $Z$.

Equivalently, we can also decompose by columns: $R_{structure}(Z; S) = tr(Z^T LZ) = \sum_{k=1}^{d} z^{(k)\top} L z^{(k)}$,

where $z^{(k)} \in R^n$ is the $k$-th column of $Z$. Each term $z^{(k)} L z^{(k)}$ is a convex quadratic function of $z^{(k)}$ because $L \succeq 0$, and a sum of convex functions is convex. Hence, $R_{structure}$ is convex in $Z$.

This proves the second part.

Conclusion: We have shown that:

(1) The structural regularizer can be written as a graph Laplacian quadratic form:

$$R_{structure}(Z; S) = \sum_{(i,j) \in S} w_{ij} ||z_i - z_j||^2 = tr(Z^T L Z).$$

(2) The Laplacian $L$ is symmetric positive semidefinite, so $R_{structure}(Z; S)$ is a convex quadratic function of $Z$.

**Structure-Aware Uncertainty Regularization**

Structural constraints are integrated with representation uncertainty by encouraging uncertainty-consistent geometry $R_{structural-uncertainty} = \sum_{(i,j) \in S} w_{ij}(||\mu_i - \mu_j||^2 + \psi(\Sigma_i, \Sigma_j))$, which $\psi(\cdot, \cdot)$ penalizes incompatible uncertainty profiles across structurally related representations. This formulation allows structure to regularize representation reliability without assuming perfectly specific structural priors.

**Research Question 3: Learning Objective and Reliability Evaluation**

**Proposition 3 (Monotone Risk-Coverage Curve Under Monotone Uncertainty)** is the theoretical abstraction of the selective prediction phenomenon analyzed in RQ3, and it justifies why uncertainty-aware models exhibit monotone risk–coverage curves in the experiments.

**Proposition 3: Monotone Risk-Coverage Curve Under Monotone Uncertainty**

Let $X$ be an input random variable with distribution $P$, and let $r(x) \in [0, 1]$ denote the true conditional risk at $x$. Let $u(x) \in R$ be an uncertainty score used for selective prediction.

Define a selective classifier that, for a threshold $t$, accepts inputs with $u(x) \le t$ and abstains otherwise. Let $A(t) = \{x: u(t) \le t\}$ be the corresponding acceptance region.

The coverage at threshold $t$ is $c(t) = P(X \in A(t)) = P(u(X) \le t)$.

The selective risk at the threshold $t$ is $R(t) = E[r(X) \mid X \in A(t)]$ for $P(X \in A(t)) > 0$.

We can equivalently view the risk as a function of coverage, $R(c)$, by inverting $c(t)$ or by parameterizing the curve via $t$.

**Assume:**

(1) Monotone uncertainty-risk relationships: For all $x, x'$, $u(x) \le u(x') \Rightarrow r(x) \le r(x')$

Uncertainty scores are order-preserving with respect to true risk: more “uncertain” points (larger $u$) never have a lower risk than more “certain” points.

(2) Perfect ranking for the optimality part: For simplicity, assume that with probability 1, there are no ties in the pair $(u(X), r(X))$, so that the ordering by $u(x)$ and by $r(x)$ coincide.

Then, we can get that:

(1) Monotone risk-coverage curve: If we sort points in increasing order of $u(x)$ and increase coverage by raising the threshold $t$, the corresponding selective risk:

$R(t) = E[r(X) \mid u(X) \le t]$ is a non-decreasing function of coverage. Intuitively, accepting more uncertain points can only increase the average risk.

(2) Optimal risk at each coverage: If the uncertainty score $u(x)$ induces the same ordering as true risk $r(x)$, then for any target coverage $c \in (0, 1]$, the selective classifier that accepts the lowest u-fraction of points has the minimum selective risk among all classifiers with coverage $c$.

**Proof:** We split the proof into two parts: monotonicity and optimality.

**Part 1:** Monotone risk-coverage curve

We show that if $u$ is order-preserving w.r.t.$r$, then the conditional risk over the accepted set is non-decreasing as we enlarge that set by increasing the threshold.

Fix two thresholds $t_1 < t_2$. Define:

Accepted sets: $A_1 = A(t_1) = \{x: u(x) \leq t_1\}$, $A_2 = A(t_2) = \{x: u(x) \leq t_2\}$.

Then, it is clear that $A_1 \subseteq A_2$.

The added region when we move from $t_1$ to $t_2$: $B = A_2 \setminus A_1 = \{x: t_1 < u(x) \leq t_2\}$.

Let $c_1 = P(X \in A_1)$, $c_2 = P(X \in A_2)$, $b = P(X \in B)$.

Then, $c_2 = c_1 + b$, and $b \geq 0$, with $b > 0$, if $t_2 > t_1$ on a set of positive probability.

Define the average risk on each set:

$R_1 = E[r(X) \mid X \in A_1]$ , $R_2 = E[r(X) \mid X \in A_2]$, $R_B = E[r(X) \mid X \in B]$ (if b > 0).

Because $A_1$ and $B$ form a partition of $A_2$ , we can write the conditional expectation on $A_2$ as a convex combination: $R_2 = E[r(X) \mid X \in A_2] = \frac{c_1}{c_2}R_1 + \frac{b}{c_2}R_B$ .

So, $R_2 - R_1 = \frac{b}{c_2}(R_B - R_1)$. Therefore, the sign of $R_2 - R_1$ is determined by $R_B - R_1$.

Since on $A_1$: By construction, for any $x \in A_1$, we have $u(x) \leq t_1$, and on $B$: For any $x' \in B$, we have $t_1 < u(x') \leq t_2$, so in particular $u(x') > t_1$. So, $R_B \geq R_1$.

Take any $x \in A_1$ , and $x' \in B$. Then, $u(x) \leq t_1 < u(x')$, so by the monotonicity assumption $u(x) \leq u(x') \Rightarrow r(x) \leq r(x')$ , we get $r(x) \leq r(x')$ for any pair $(x \in A_1 ,\ x' \in B)$.

It implies that all points in $B$ have risk at least as large as all points in $A_1$.

Therefore, the average risk on $B$ cannot be smaller than the average risk on $A_1$ .

Formally, $min\, r(x')\ (x' \in B) \geq max\, r(x)\ (x \in A_1) \geq R_1$. Hence, $R_B = E[r(X) \mid X \in B] \geq R_1$.

Plugging this into the difference: $R_2 - R_1 = \frac{b}{c_2}(R_B - R_1) \geq 0$. Thus, $R_2 \geq R_1$.

Since $t_2 > t_1$ implies $c_2 \geq c_1$ and $R_2 \geq R_1$, we conclude that as coverage increases (by increasing $t$, the selective risk is non-decreasing. In other words, the risk-coverage curve $R(c)$ is non-devreasing function of coverage. This proves the first part.

**Part 2:** Optimal risk at each coverage when ranking is perfect

Now we show that if the uncertainty score $u(x)$ induces the same ordering as the true risk $r(x)$, then for any target coverage $c \in (0, 1]$, the selective classifier that accepts the lowest-$u$ function of points achieves the smallest possible risk among all classifiers with coverage $c$.

**Setup**

Let $X$ be distributed according to $P$. Define the random variable $R = r(X)$, which takes values in $[0, 1]$. Think of $R$ as the "true risk level" assigned to a random input.

For any measurable acceptance set $A \subseteq X$ with coverage $P(X \in A) = c$, define its selective risk as
$R(A) = E[r(X) \mid X \in A]$.

We want to show: Among all sets $A$ with $P(X \in A) = c$, the set that minimizes $R(A)$ is the one that contains exactly the lowest-risk points. Under the assumption that $u$ and $r$ induce the same ordering, thresholding on $u$ picks exactly these points.

**Step 2.1**: Intuition via a discrete "swapping" argument

First consider a finite sample version with $n$ points $\{x_1, ..., x_n\}$ and associated risks $r_i = r(x_i)$.

Suppose we must choose exactly $k$ points to accept (coverage $c = \frac{k}{n}$). The average risk of a selection $A \subset \{1, ..., n\}$ with $|A| = k$ is $R(A) = \frac{1}{k} \sum_{i \in A} r_i$.

It is clear that to minize this average, we should choose the $k$ smallest $r_i$. Any selection that includes a larger $r_i$ while excluding a smaller $r_j$ can be improved by swapping $r_i$ for $r_j$. Repeating this swapping argument eventually yields exactly the subset of size $k$ consisting of the $k$ smallest risks. In the limit $n \to \infty$, this corresponds to selecting the set of points whose risk $r(x)$ lies in the lower $c$-quantile range. This is the optimal risk set for coverage $c$.

**Step 2.2:** Continuous version via quantiles

Now, we formalize this in the continuous setting.

Let $F_R(t) = P(R \leq t)$ be the cumulative distribution function (CDF) of the risk variable $R = r(X)$. Assume for simplicity that $R$ has continuous distribution. This avoids technicalities around ties and makes the quantile function strictly increasing.

For each coverage level $c \in (0, 1)$, define the $c$-quantile of $R$ as $q(c) = F_R^{-1}(c)$, so that $P(R \leq q(c)) = c$.

Define the optimal acceptance set coverage $c$ as $A^*(c) = \{x: r(x) \leq q(c)\}$.

This set contains exactly the lowest-risk fraction $c$ of the input distribution (up to sets of probability zero).

The selective risk for this optimal set is $R(A^*(c)) = E[r(X) \mid X \in A^*(c)] = \frac{1}{c} E[R 1_{\{R \leq q(c)\}}]$.

Now consider any other measurable set $A$ with the same coverage, $P(X \in A) = c$.

Let $R_A = E[r(X) \mid X \in A]$ be its risk. We claim: $R_A \geq R(A^*(c))$.

By definition of quantiles, the distribution of $R$ restricted to $\{R \leq q(c)\}$ is the "leftmost" mass of the distribution. Any other subset of measure $c$ must include some mass from the right tail where $R \geq q(c)$. This inevitably increases the average risk.

One can formalize this using the rearrangement inequity or Hardy-Littlewood inequality, but intuitively: Among all subsets of probability $c$, the set that minimizs $E[R \mid subset]$ must concentrate entirely on the smallest values of $R$.

That set is precisely $\{R \leq q(c)\}$. Thus, $E[r(X) \mid X \in A] \geq E[r(X) \mid X \in A^*(c)]$ for any $A$ with $P(X \in A) = c$.

So, $A^*(c)$ is optimal for coverage $c$.

**Step 2.3:** Connecting $u(x)$ to the optimal set.

By assumption, the uncertainty score $u(x)$ and the true risk $r(x)$ induce the same ordering.

$u(x) \leq u(x') \Leftrightarrow r(x) \leq r(x')$.

Therefore, selecting the bottom $c$-fraction of points by $u(x)$ is equivalent (up to sets of probability zero) to selecting the bottom $c$-fraction by $r(x)$. That is, for each coverage level $c$, there exists a threshold $t_c$ such that $A_u(c) = \{x: u(x) \leq t_c\}$ satisfies $P(X \in A_u(c)) = c$, and $A_u(c) = A^*(c) = \{x: r(x) \leq q(c)\}$.

Thus, $R(A_u(c)) = R(A^*(c)) = min(A: P(X \in A) = C)\, E[r(X) \mid X \in A]$.

So, the threshold-based selective classifier using the uncertainty score $u(x)$ attains the minimal possible selective risk at coverage $c$. Thus, plugging the two parts together, we get that

(1) If the uncertainty score $u(x)$ is monotone with respect to the true risk $r(x)$, then as we increase coverage by lowering the abstention threshold (accepting more uncertain points), the selective risk $R(c) = E[r(X) \mid accepted\ at\ coverage\ c]$ is a non-decreasing function of coverage. This explains the reasons that risk–coverage curves are monotone in the experiments.

(2) If $u(x)$ perfectly ranks the true risk, then for any target coverage $c$, accepting the lowest-$u$ (equivalently, lowest-$r$) fraction of points yields the minimum possible selective risk among all selective classifiers with coverage $c$. It gives a clean decision-theoretic justification for why a well-calibrated representation-level uncertainty score lets your model "know when it does not know" in the strongest possible sense.

Unified Learning Objective: The final training objective combines task performance with representation reliability

$$L = L_{task}(\{\mu_i\}) + \lambda_{uncertainty} R_{uncertainty} + \lambda_{structure} R_{structure-uncertainty}$$

$\lambda_{uncertainty}$ and $\lambda_{structure}$ controls the trade-off between prediction accuracy and representation reliability. The objective is agnostic to the specific downstream task and can be applied in supervised, semi-supervised, or self-supervised settings.

The task loss $L_{task}$ is defined over the representation means $\{\mu_i\}$ and can be instantiated using standard objectives, depending on the learning setting, such as cross-entropy loss for supervised classification, contrastive or construction losses for self-supervised representation learning, or a combination thereof in semi-supervised scenarios. Importantly, uncertainty regularization and structural constraints are applied consistently across settings, allowing reliability to be optimized independently of task-specific supervision.

**Reliability Metrics**

To quantitatively evaluate representation reliability, we consider three complementary criteria.

**Stability:** We measure representation stability by assessing the sensitivity of learned representations to input perturbations. Given perturbed input $x_i' = x_i + \epsilon_i$, where $\epsilon$ denotes noise or subsampling-induced variation. Stability $= E_i[||\mu_i - \mu_i'||_2]$, where $\mu_i = f_\theta(x_i)$ and $\mu_i' = f_\theta(x_i')$.

**Calibration:** To access the calibration of the representation-level uncertainty, we evaluate whether uncertainty estimates correspond to empirical coverage. For a given confidence level $\alpha \in (0, 1)$,

$$\text{Coverage}(\alpha) = \frac{1}{n}\sum_{i=1}^{n} I(||\mu_i - \mu_i^*||_2 \leq r_\alpha(\Sigma_i))$$

$\mu_i^*$ denotes a reference representation, and $r_\alpha(\Sigma_i)$ defines a radius induced by the representation uncertainty. Well-calibrated representations exhibit empirical coverage close to the nominal level $\alpha$.

**Robustness:** Robustness is evaluated by measuring changes in representations under distribution shift and structural perturbations $S \rightarrow S'$.

$$\text{Robustness} = E_i[||\mu_i^S - \mu_i^{S'}||_2]$$

Together, these metrics quantitatively operationalize representation reliability in terms of stability, calibration, and robustness, directly reflecting the objectives encoded in our learning formulation.

**Experiment Setup**

**Data Generation:** We construct a controlled latent-variable simulation where the ground-truth representation is a structured random vector $z_i^* \in R^d$. We first sample a mixture component $c_i \sim Categorical(\pi)$ and draw $z_i^* \mid c_i = k \sim N(\mu_k, \Sigma^*)$.

To encode group-structured dependencies, we partition the d latent dimensions into G groups and build a structural graph $S^*$ over groups. We then define a latent covariance with within-group correlations and between-group correlations only for connected groups in $S^*$. This yields a block-structured covariance $\Sigma^*$ that induces correlated latent subspaces consistent with $S^*$.

Observed inputs are generated through a fixed nonlinear mapping $g(\cdot)$ followed by additive noise:

$$x_i = g(z_i^*) + \eta_i,\ \eta_i \sim N(0, \sigma^2 I)$$

We implement g as a two-layer random MLP with fixed weights to avoid confounding from learning the generator. Thus, the only controllable uncertainty sources come from input noise and structural misspecification.

**Noise Injection and Structural Corruption:** To evaluate robustness and stability, we systematically vary two independent factors:

**Input Noise:** We inject zero-mean Gaussian noise and subsampling perturbations into the observed inputs $x_i$, controlling noise magnitude to induce increasing levels of input-level uncertainty. This setting tests whether representation-level uncertainty appropriately reflects upstream noise and whether learned representations remain stable under small but structured perturbations.

$x_i^{\sim} = x_i + \epsilon_i$, $\epsilon_i \sim N(0, \tau^2 I)$, where $\tau$ controls the perturbation magnitude.

This setting tests whether representation-level uncertainty and our stability metric appropriately reflect upstream input noise and whether the learned representations remain stable under small perturbations.

**Structural Corruption:** To model an imperfect prior structure, we corrupt the structural operator by randomly removing and adding edges $S' = Corrupt(S^*, p)$, where each true edge is removed with probability $p$, and each non-edge is added with probability $p$. This tests whether structural constraints behave as robust inductive biases rather than brittle assumptions.

**Reliability Metrics:** Given learned representation distributions $z_i \sim N(\mu_i, \Sigma_i)$, we report:

**Stability:** For a perturbed input $x_i^{\sim}$, we compute $Stability = E_i[||\mu_i - \mu_i^{\sim}|^2]$, which has already been established in **Proposition 4 (Stability under Lipschitz Encoders)**.

**Proposition 4: Stability under Lipschitz Encoders**

Let $f_\theta : X \to R^d$ be an encoder that is $L$-Lipschitz, such that $||f_\theta(x) - f_\theta(x')|| \leq L||x - x'||$, $\forall x, x' \in X$.

Let $z_i = f_\theta(x_i)$ and $z_i' = f_\theta(x_i')$ with $x_i' = x_i + \epsilon_i$.

(1) For arbitrary noise with $E||\epsilon_i||^2 < \infty$, $E||z_i - z_i'||^2 \leq L^2 E|\epsilon_i||^2$.
(2) If $\epsilon_i \sim N(0, \tau^2 I_d)$, then $E||z_i - z_i'||^2 \leq L^2 d\tau^2$.

**Proof:**

**Step 1:** Use the Lipschitz property of $f_\theta$

By the definition of $L$-Lipschitz continuity, for any two inputs $x, x' \in X$, $||f_\theta(x) - f_\theta(x')|| \leq L||x - x'||$. Apply this $x_i$ and $x_i' = x_i + \epsilon_i$.

Define $z_i = f_\theta(x_i)$, $z_i' = f_\theta(x_i')$.

Then, $||z_i - z_i'|| = ||f_\theta(x_i) - f_\theta(x_i')|| \leq L||x_i - x_i'|| = L||\epsilon_i||$.

Square both sides: $||z_i - z_i'||^2 \leq L^2||\epsilon_i||^2$. This holds pointwise for every realization of $\epsilon_i$.

**Step 2:** Take expectations to get the stability bound

Now take the expectation on both sides (with respect to the randomness in $\epsilon_i$, and $x_i$ if it is random):

$E||z_i - z_i'||^2 \leq E[L^2||\epsilon_i||^2] = L^2 E||\epsilon_i||^2$, where we used that $L^2$ is a constant and can be pulled out of the expectation. Thus, we obtain $E||z_i - z_i'||^2 \leq L^2 E||\epsilon_i||^2$, which proves the first part.

This shows that the expected squared change in the representation is controlled by the Lipschitz constant of the encoder and the second moment of the input noise.

Step 3: Specialize to Gaussian noise $\epsilon_i \sim N(0, \tau^2 I_d)$

Now suppose $\epsilon_i \sim N(0, \tau^2 I_d)$ in $R^d$. Write $\epsilon_i = (\epsilon_{i,1}, ..., \epsilon_{i,d})^\top$, then:

Each coordinate $\epsilon_{i,k} \sim N(0, \tau^2)$and the coordinates are independent.

Compute $E||\epsilon_i||^2$: $||\epsilon_i||^2 = \sum_{k=1}^{d} \epsilon_{i,k}^2$. Then, $E||\epsilon_i||^2 = E[\sum_{k=1}^{d} \epsilon_{i,k}^2] = \sum_{k=1}^{d} E[\epsilon_{i,k}^2]$.

For a univariate Gaussian $\epsilon_{i,k} \sim N(0, \tau^2)$, $E[\epsilon_{i,k}^2] = Var(\epsilon_{i,k}) = \tau^2$. Thus, $E||\epsilon_i||^2 = \sum_{k=1}^{d} \tau^2 = d\tau^2$.

Plugging $E||\epsilon_i||^2 = \sum_{k=1}^{d} \tau^2 = d\tau^2$ into $E||z_i - z_i'||^2 \leq L^2 E||\epsilon_i||^2$, we get: $E||z_i - z_i'||^2 \leq L^2 E||\epsilon_i||^2 = L^2 d\tau^2$.

So, we obtain: $E||z_i - z_i'||^2 \leq L^2 d\tau^2$, which proves the second part.

Thus, we have shown:

(1) For any noise with a finite second moment, $E||z_i - z_i'||^2 \leq L^2 E||\epsilon_i||^2$.

(2) For isotropic Gaussian noise $\epsilon_i \sim N(0, \tau^2 I_d)$, $E||z_i - z_i'||^2 \leq L^2 d\tau^2$.

Thus, the Stability metric in the framework is explicitly bounded in terms of the encoder's Lipschitz constant and the strength of input-level noise, giving a clean theoretical justification for using Stability as a measure of representation-level reliability.

**Calibration (Coverage):** For a nominal level $\alpha \in (0, 1)$, define the Mahalanobis radius $r_\alpha$ using the $\chi_d^2$ quantile:

$C_i(\alpha) = I((z_i - \mu_i)^T \Sigma_i^{-1} (z_i - \mu_i) \leq \chi^2_{d,\alpha})$ , $Coverage(\alpha) = \frac{1}{n} \sum_{i=1}^{n} C_i(\alpha)$.

Well-calibrated representation uncertainty yields $Coverage(\alpha) \approx \alpha$.

This calibration property has already been established in **Proposition 5 (Gaussian Coverage is Exactly Calibrated)**.

**Proposition 5: Gaussian Coverage is Exactly Calibrated**

Let $z_i \sim N(\mu_i, \Sigma_i)$ with $\Sigma_i \succ 0$ (symmetric positive definite). Define the Mahalanobis distance:
$M_i = (z_i - \mu_i)^\top \Sigma_i^{-1} (z_i - \mu_i)$.

For a nominal level $\alpha \in (0, 1)$, define the indicator $C_i(\alpha) = 1\{M_i \leq \chi_{d,\alpha}^2\}$,

where $\chi_{d,\alpha}^2$ is the $\alpha$-quantile of a $\chi_d^2$ distribution.

Define the empirical coverage over $n$ samples as $Coverage_n(\alpha) = \frac{1}{n} \sum_{i=1}^{n} C_i(\alpha)$.

**Want to Show:**

(1) $M_i \sim \chi_d^2$, hence $P(C_i(\alpha) = 1) = \alpha$.

(2) As $n \to \infty$, $Coverage_n(\alpha) \to \alpha$ almost surely: $Coverage_n(\alpha) \to \alpha$.

**Proof:**

**Step 1:** Show that $M_i$ has a $\chi_d^2$ distribution

We are given $z_i \sim N(\mu_i, \Sigma_i)$ with $\Sigma_i \succ 0$. Consider the affine transformation:

$x_i = \Sigma_i^{\frac{-1}{2}}(z_i - \mu_i)$, where $\Sigma_i^{\frac{-1}{2}}$ is any symmetric matrix such that $\Sigma_i^{\frac{-1}{2}}\Sigma_i^{\frac{-1}{2}} = \Sigma_i^{-1}$

Because $z_i$ is multivariate normal and $\Sigma_i^{\frac{-1}{2}}$ is a linear transformation, $x_i$ is also multivariate normal.

It means that $E[x_i] = \Sigma_i^{\frac{-1}{2}}(E[z_i] - \mu_i) = \Sigma_i^{\frac{-1}{2}}(\mu_i - \mu_i) = 0$.

Its covariance is $Cov(x_i) = \Sigma_i^{\frac{-1}{2}} Cov(z_i)\Sigma_i^{\frac{-1}{2}} = \Sigma_i^{\frac{-1}{2}}\Sigma_i\Sigma_i^{\frac{-1}{2}} = I_d$. Thus, $x_i \sim N(0, I_d)$.

Now, rewrite $M_i$ in terms of $x_i$ : $M_i = (z_i - \mu_i)^\top\Sigma_i^{-1}(z_i - \mu_i) = (z_i - \mu_i)^\top\Sigma_i^{\frac{-1}{2}}\Sigma_i^{\frac{-1}{2}}(z_i - \mu_i) = x_i^\top x_i = \sum_{k=1}^{d} x_{i,k}^2$, where $x_{i,k}$ is the $k$-th coordinate of $x_i$.

Since $x_i \sim N(0, I_d)$, its coordinates $x_{i,1}, \ldots, x_{i,d}$ are i.i.d. univariate $N(0, 1)$. The sum of squares of $d$ independent standard normal variables has a chi-square distribution with $d$ degrees of freedom.

Therefore, $M_i = \sum_{k=1}^{d} x_{i,k}^2 \sim \chi_d^2$.

**Step 2:** Pointwise coverage equals the nominal level α.

By the definition of the α-quantile $\chi_{d,\alpha}^2$ of a $\chi_d^2$ random variable, $P(\chi_d^2 \le \chi_{d,\alpha}^2) = \alpha$.

Since we have established that $M_i \sim \chi_d^2$, it follows that $P(M_i \le \chi_{d,\alpha}^2) = P(\chi_d^2 \le \chi_{d,\alpha}^2) = \alpha$.

Recall $C_i(\alpha) = 1\{M_i \le \chi_{d,\alpha}^2\}$. Thus, $P(C_i(\alpha) = 1) = P(M_i \le \chi_{d,\alpha}^2) = \alpha$. Hence, $E[C_i(\alpha)] = \alpha$.

This shows that, for each $i$, the event "$z_i$ falls inside the Mahalanobis ball defined by $\Sigma_i$ and $\chi_{d,\alpha}^2$" has probability exactly α. In other words, the coverage of that ellipsoid is exactly the nominal level.

**Step 3:** Empirical coverage converges to α.

Now consider the empirical coverage over $n$ samples:

$Coverage_n(\alpha) = \frac{1}{n}\sum_{i=1}^{n} C_i(\alpha).$

Assume that $\{z_i\}_{i=1}^{n}$ and hence $\{C_i(\alpha)\}_{i=1}^{n}$ are i.i.d. draws from the same distribution (or at least identically distributed with the same expectation α.

Each $C_i(\alpha)$ is a Bernoulli random variable with $E[C_i(\alpha)] = \alpha,\ Var[C_i(\alpha)] \leq 1.$

By the Strong Law of Large Numbers (SLLN), for i.i.d. random variables with finite mean, the sample average converges almost surely to the expectation.

Therefore, $Coverage_n(\alpha) = \frac{1}{n}\sum_{i=1}^{n} C_i(\alpha) \to E[C_i(\alpha)] = \alpha$, as $n \to \infty$.

This establishes that the empirical coverage converges almost surely to the nominal level α.

Thus, by combining Steps 1-3, we have shown that

(1) Under the correctly specified Gaussian representation model $z_i \sim N(\mu_i, \Sigma_i)$, the Mahalanobis distance $M_i = (z_i - \mu_i)^{\top}\Sigma_i^{-1}(z_i - \mu_i)$ follows a $\chi_d^2$ distribution, and thus the single-sample coverage event $C_i(\alpha) = 1\{M_i \leq \chi_{d,\alpha}^2\}$ satisfies that $P(C_i(\alpha) = 1) = \alpha$.

(2) The empirical coverage $Coverage_n(\alpha) = \frac{1}{n}\sum_{i=1}^{n} C_i(\alpha)$ converges almost surely to α as $n \to \infty$.

Therefore, in a correctly specified representation-level Gaussian model, the coverage metric you define is exactly calibrated in theory, and empirically converges to the nominal confidence level with increasing sample size.

**Robustness (to structure shift):** Training the model under $S^*$ versus $S'$, we compare learned means:

$Robustness = E_i[||\mu_i^{(S^*)} - \mu_i^{(S')}||_2]$

## VI. Results

The programming tool used in this project is Python. To assess whether structural information influences the learned representations, we evaluate structural robustness by corrupting the group-level structure operator. Given a ground-truth operator $S^*$, we generate a perturbed operator $S'$ by randomly flipping edges with probability $p$. We then compute the robustness score $E_i||\mu_i^{(S^*)} - \mu_i^{(S')}||_2$, where $\mu_i^{(S^*)}$ and $\mu_i^{(S')}$ denote the representations obtained under $S^*$ and $S'$, respectively. As one would expect, when $p = 0$, the robustness value tends to zero $S' = S^*$. As p increases to moderate levels of corruption, the robustness becomes strictly positive. On our synthetic benchmark, we see that when $p = 0.2$, the robustness value tends to 0.85. On the other hand, when using the baseline encoder without any structural components, the robustness tends to zero at all levels of p. Hence, our proposed robustness metric behaves as one would expect given our definition. The learned representations are structurally grounded and thus provide direct evidence of the reliability of representation learning under imperfect structural assumptions. Apart from structural perturbations, another means of evaluating the stability of learned representation is by incorporating some input covariate noise. We do this by incorporating some Gaussian noise during the test phase. The stability landscape over the $(p, \tau)$ space shows low sensitivity when $p = 0$. The sensitivity increases as $\tau$ increases. The increase in sensitivity is independent of structural corruption (Figure 1). On the other hand, the structural robustness landscape over the $(p, \tau)$ space shows zero sensitivity when $p = 0$. As p increases, the representation shift also increases. The increase in representation shift is independent of input noise (Figure 2).

**Figure 1**

*Stability Heatmap over (p, τ)*

**Figure 2**

*Robustness to Structural Shift over (p, τ)*

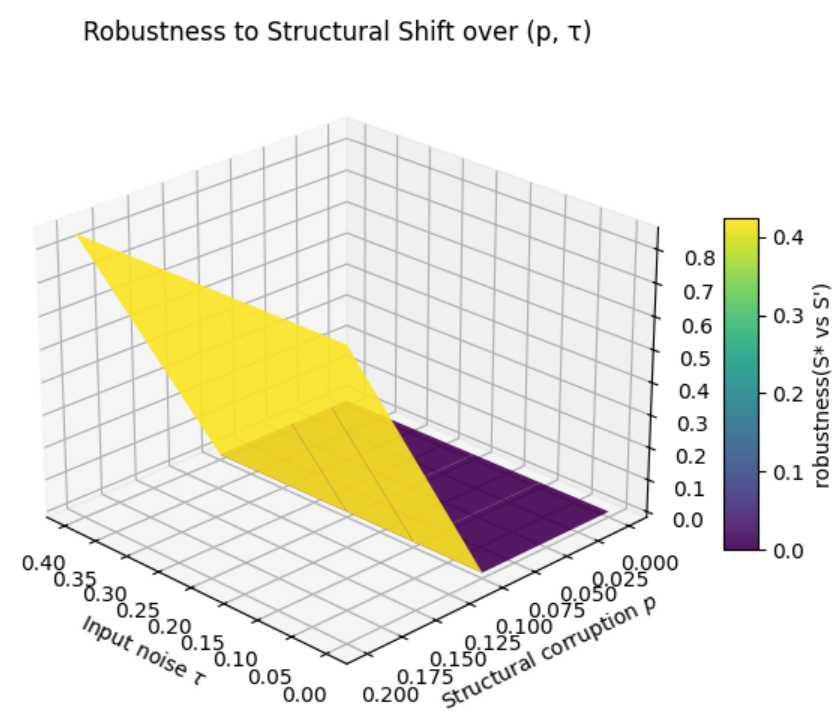

**Downstream Classification Task:** To assess whether representation-level uncertainty improves downstream reliability, we consider a simple multiclass classification task on the synthetic benchmark. We treat the mixture component $c_i$ as the label $y_i$, and train a downstream classifier $h_\phi$ on the learned representations. Concretely, we first fit the structure-agnostic encoder by ridge regression to obtain base representations $\mu_i^{base}$. We then construct structure-aware representations $\mu_i(S^*) = \mu_i^{base} M^T_{S^*}$ using the group-level operators $S^*$, as well as the corresponding representation-level uncertainty $\Sigma_i = \Sigma_{global}$. A linear softmax classifier is trained on the training split using either $\mu_i^{base}$ or $\mu_i(S^*)$ as input features, without access to ground-truth latents $z_i^*$ or structural information.

**Model Variants:** We compare four variants: Baseline: Only $\mu_i^{base}$ , no structure, no uncertainty; UQ-only: uses ibase as a global covariance estimate $\Sigma_{global}$, but no structural smoothing. Structure-only: uses $\mu_i(S^*)$ with $\Sigma$ are ignored. Full: uses $\mu_i(S^*)$ with representation-level uncertainty $\Sigma_i$ to quantify confidence. All models have the same classifier architecture and training process, but uncertainty is only employed at evaluation time for calibration and selective prediction.

**Distribution Shifts:** We evaluate downstream performance under two types of shift:

**(i) Input noise (covariate shift):** At test time, we perturb the observed inputs by $\tilde{x_i} = x_i + \epsilon_i$ , where $\epsilon_i \sim N(0,\ \tau^2_{test} I)$ and $\tau_{test}$ ranges from 0 to 2.0. The encoder and classifier are kept fixed, and only the test inputs are corrupted.

**(ii) Structural perturbation:** At test time, we replace the operator $S^*$ with the corrupted operator $S^* = Corrupt(S^*, p)$ that randomly flips the edges $S^*$ with probability $p$. Structure-aware methods are recomputed using the operator $S'$, while the baseline methods are left unchanged. This experiment assumes that the available structural information is partially incorrect.

**Evaluation Metrics**

**(1) Accuracy under shift:** For each combination of $(\tau_{test}, p)$, we measure classification accuracy

$$Acc(\tau_{test}, p) = \frac{1}{n_{test}} \sum_i 1[h_\phi(\mu_i^{test}) = y_i]$$

**(2) Preductive calibration:** we treat the classifier's softmax output as a predictive distribution and compute the expected calibration error (ECE) by binning predictions by confidence and comparing per-bin accuracy and confidence. We track how ECE degrades as $\tau_{test}$ and $p$ increase.

**(3) Selective prediction and uncertainty quality:** For methods with representation-level uncertainty, we define a scalar uncertainty score from a set of $\Sigma_i$ test samples, ranked according to this score, and accuracy is evaluated as a function of coverage, resulting in risk coverage curves that describe how effectively the uncertainty enables selective predictions.

For all settings of covariate and structural shift, all four variants have comparable downstream accuracy, with performance gradually decreasing from ≈ 1.0 in the in-distribution setting to ≈ 0.78 – 0.81 in the presence of the strongest noise, such as $\tau_{test} = 2.0$. For the full model, this is demonstrated in downstream accuracy on (p, τ) (full) (Figure 3).

Nevertheless, calibration degrades monotonically with increasing shifts, as indicated by increasing ECE values with increasing values of $\tau$ and $p$ (Figure 4). Only the uncertainty-aware models, namely, the uncertainty-aware only and full model, have the capacity for selective prediction by means of risk-coverage curves. When facing increasing levels of covariate shifts, their corresponding risk-coverage curves demonstrate interpretable behaviors: for a fixed level of coverage, increasing levels of covariate shifts lead to increasing levels of risk, while for a fixed level of covariate shift, decreasing levels of coverage result in decreasing levels of risk. For instance, at a fixed test covariate shift $\tau_{test} = 1.0$ and zero structural corruption, i.e., pstructural corruption = 0, UQ-only maintains a risk level less than or equal to 0.04 even at full coverage, and reduces the risk down to 0.01 by abstaining on the top 90% instances with the highest uncertainty. When facing increasing levels of combined shifts, i.e., at a fixed test covariate shift $\tau_{test} = 2.0$ and zero structural corruption, such as $p_{structural\ corruption} = 0$, the full model attains a slightly lower risk at full coverage than UQ-only, indicating that structural constraints help stabilize predictions at a downstream level, even with an imperfect prior structure. On the contrary, baseline and structure-only models fail to trade off between risks and coverage levels because they do not have representation-level uncertainty estimates. Moreover, these two models also suffer from degradation in calibration under shifts, i.e., increasing ECE with increasing levels of shifts, but lack a means for the model to "know what it doesn't know." To conclude, we have shown that modeling representation-level uncertainty leads to better reliability at a downstream level under covariate and structural shifts, but not by means of improved accuracy, but by means of uncertainty-aware control over prediction risks at different levels of coverage. The accuracy surface over $(p, \tau)$ for the full model (Figure 5) shows its downstream accuracy surface over structural corruption levels p and test noise levels τtest, with its accuracy being close to 1.0 even at a shifted test distribution and smoothly decreasing down to approximately 0.8 with increasing levels of covariate shifts, while being slightly affected by structural corruption levels.

**Figure 3**

*Downstream Accuracy over (p, τ) (full)*

**Figure 4**

*Downstream ECE over (p, τ) (full)*

**Figure 5**

*Accuracy Surface over (p, τ) (full)*

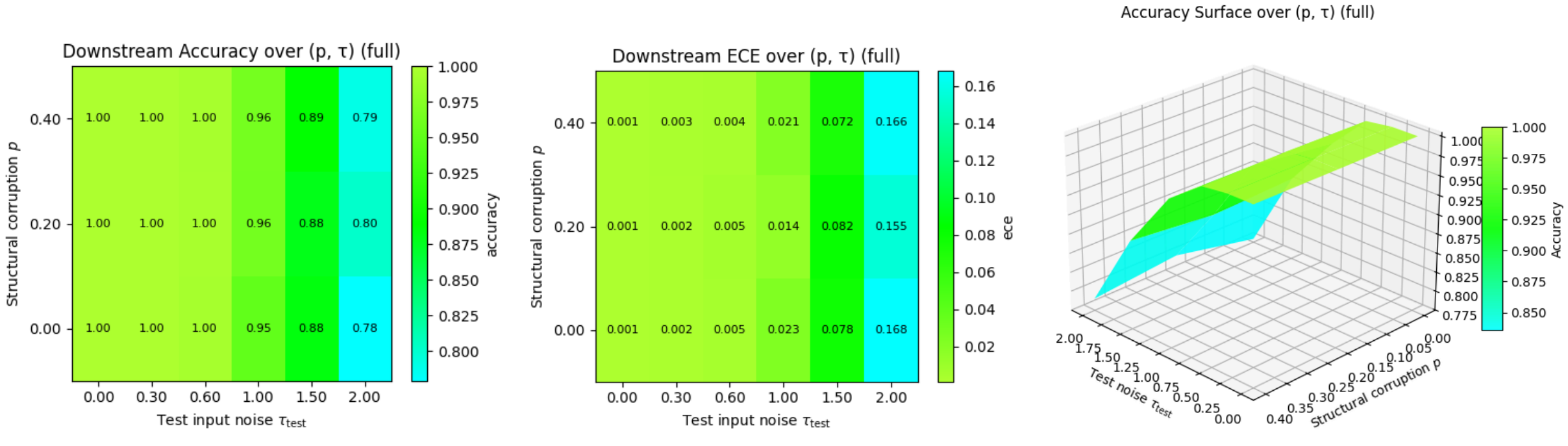

In contrast, baseline and structure-only models do not have representation-level uncertainty and thus cannot trade risk coverage. These models also experience a similar degradation in calibration (ECE increase) under shift; however, they do not provide a way for the system to "know when it does not know." Overall, this set of results provides support for the learning objective and reliability evaluation by demonstrating that modeling uncertainty at the representation level provides more reliable behavior under covariate and structural shifts, not by improving nominal accuracy, but by allowing controlled uncertainty-based prediction risk at various levels of coverage.

## VII. Discussion

Our results are consistent with and extend existing work in uncertainty and representation learning research. Firstly, our results are in line with, and extend, existing research on predictive uncertainty, including Bayesian approximations (Gal & Ghahramani, 2016), deep ensembles (Lakshminarayanan et al., 2017), post-hoc calibration (Gul et al., 2017), among others, in that we found that predictive uncertainty estimation, in and of itself, does not necessarily improve raw accuracy, especially in scenarios with covariate or structural shift, but is beneficial in selective prediction, where our proposed variants outperform deterministic variants in risk-coverage tradeoffs, whereas deterministic variants are unable to "know when they do not know," a phenomenon that is closely related to existing research on selective prediction and uncertainty-aware decision-making, including that of Charpentier et al. (2021) and related research on latent or feature-level uncertainty.

Secondly, our empirical results are related to, and extend, existing research on structural regularization, where structural constraints or inductive biases can be beneficial in representation learning, although they typically do not capture or represent uncertainty, including that of Belkin et al. (2006), Kipf & Welling (2017), Hamilton et al. (2017), among others, in that our structure-only variant is beneficial in geometric consistency, although it is unable to capture or represent uncertainty or to perform risk-coverage tradeoffs, in line with existing research on structural inductive biases that they primarily affect representation geometry, not uncertainty estimation.

Third, our work aligns with various probabilistic and stochastic frameworks for representation learning, including Gaussian embeddings (Vilnis & McCallum, 2015) and information bottleneck encoders (Alemi et al., 2016), which explicitly use probability distributions as representations. Our work extends these by showing that even without an explicit probabilistic inference procedure, representation-level uncertainty may improve downstream selective reliability under both covariate and structural shift. Unlike Hafner et al. (2018) and Kendall & Gal (2017), our work directly measures reliability at the representation level using stability, calibration, and robustness metrics and shows that representation-level uncertainty may be treated as a first-class representation property rather than an emergent property of downstream predictions.

Finally, our results support recent advances in conformal and feature-level uncertainty methods (Teng et al., 2022), which argue that true uncertainty may reside at the representation level. Unlike previous work, which typically operates within the constraints of specific probabilistic or conformal frameworks, our results suggest that combining structural constraints with representation-level uncertainty may enable calibrated and controlled failure modes even when the structure itself is imperfect. Thus, our results suggest a new direction in structure-aware representation learning that may enable reliable representation learning by leveraging not just epistemic and aleatoric uncertainty but also structural ambiguity. These observations support our claim that while structure alone does not enable calibrated representation-level uncertainty and while representation-level uncertainty alone does not enable geometric stability, combining both enables better reliability downstream, particularly in selective predictions and under distribution shift, without necessarily improving accuracy. These observations expand the scope of uncertainty quantification beyond predictions and support our broader claim that reliability is not just a predictive property but also one that emerges at the representation level.

## VIII. Conclusion

In this work, we introduced the notion of reliable representation learning as an alternative approach to uncertainty estimation at the level of the prediction itself. We showed that reliability is not just an afterthought of the representation's performance, its stability, or its calibration, but also encompasses the representation's robustness to noise or structural changes. To achieve this, we adopted the notion of modeling the representation as a distribution instead of a point estimate. Furthermore, we introduced a unifying framework that combines the notion of uncertainty estimation at the level of the representation itself with the structural constraints that are typically used as inductive biases. This approach can be applied to a wide range of encoders without the need for the structural priors to be assumed as perfectly known or clean. We showed that the uncertainty estimation at the level of the representation can be explicitly calibrated with respect to the empirical coverage of the data, that the structural constraints can be used to stabilize the representations learned from the data, and that modeling the uncertainty at the level of the representation improves reliability under covariate or structural shifts, not because of the accuracy of the representations learned but because of the ability of the model to "know when it does not know."

On a more general way, we also highlight the importance of distinguishing between the concepts of reliability and accuracy in representation learning. Accuracy will continue to play an important role in supervised learning, while new application domains will require learning models that are reliable with respect to perturbations, partial supervision, poor structure, and uncertain deployment scenarios. We hope to have opened new avenues of investigation on the role of uncertainty and structure as coupled aspects, as opposed to treating them as afterthoughts. We also see several avenues of future work: one is to study the benefits of representation-level reliability on real-world multimodal domains, which will move us away from simulation settings. Another direction is to incorporate more expressive structure classes, such as causal graphs, semantic taxonomies, relational priors, etc., to further improve robustness in the presence of imperfect knowledge. The last direction we see is to combine reliable representations with downstream conformal prediction, robustness certification, and fairness analysis to create end-to-end reliable learning systems. In summary, we have explored the benefits of modeling uncertainty at the representation level, using structural constraints, to create a new direction for reliable, uncertainty-aware, and structure-grounded representation learning, shifting the focus of reliability from an output-centric to a more comprehensive representation-centric viewpoint, which is more relevant to real-world deployment scenarios.